# Evaluating the Challenges of LLMs in Real-world Medical Follow-up: A Comparative Study and An Optimized Framework

JINYAN LIU, College of Biomedical Engineering and Instrument Science, Zhejiang University, China

ZIKANGCHEN, College of Biomedical Engineering and Instrument Science, Zhejiang University, China

QINCHUAN WANG, Department of Surgical oncology, Zhejiang University School of Medicine Sir Run Run Shaw Hospital, China

TAN XIE, Department of Surgical oncology, Zhejiang University School of Medicine Sir Run Run Shaw Hospital, China

HEMING ZHENG, Department of Surgical oncology, Zhejiang University School of Medicine Sir Run Run Shaw Hospital, China

XUDONG LV[*], College of Biomedical Engineering and Instrument Science, Zhejiang University, China

When applied directly in an end-to-end manner to medical follow-up tasks, Large Language Models (LLMs) often suffer from uncontrolled dialog flow and inaccurate information extraction due to the complexity of follow-up forms. To address this limitation, we designed and compared two follow-up chatbot systems: an end-to-end LLM-based system (control group) and a modular pipeline with structured process control (experimental group). Experimental results show that while the end-to-end approach frequently fails on lengthy and complex forms, our modular method—built on task decomposition, semantic clustering, and flow management—substantially improves dialog stability and extraction accuracy. Moreover, it reduces the number of dialogue turns by 46.73% and lowers token consumption by 80% to 87.5%. These findings highlight the necessity of integrating external control mechanisms when deploying LLMs in high-stakes medical follow-up scenarios.

**ACM Reference Format:**

---

[*] Corresponding author.

# 1 INTRODUCTION

With advances in the economy and medical technology, people's expectations for health and quality of life have risen. Many patients not only seek effective treatment but also hope to maintain a high quality of life even under health limitations[11]. Regular follow-up has become an important means to improve treatment outcomes and patient well-being. Medical follow-up refers to the process in which healthcare professionals monitor and track patients' health status over time, serving as essential feedback for treatment.

Follow-up is an indispensable part of modern healthcare systems, playing a vital role not only in individual health management but also in advancing medical research and clinical education. Traditional follow-up methods include in-person visits, telephone follow-ups, and paper-based questionnaires. While in-person visits are suitable for complex cases, they are limited by high costs and patient travel burdens [7]. Telephone follow-ups, conducted by trained hospital staff, enhance communication, improve adherence, and reduce hospital readmission rates[1]. Paper forms allow for detailed reporting but face challenges in efficiency, data management, and long-term tracking.

With the development of information technologies, electronic and online follow-up methods have gained traction due to their improved efficiency, reduced labor costs, and better data handling[4, 17]. Recently, researchers have explored using dialogue-based chatbots for intelligent follow-up data collection. Compared to human interviewers, patients are often more willing to disclose truthful information when interacting with conversational agents[6]. Chatbots also offer greater interactivity and engagement than form-based methods[2], and help alleviate the workload of healthcare professionals. However, most existing systems rely heavily on large datasets for training, making them difficult to generalize across diverse follow-up scenarios. Furthermore, rule-based systems often lack the flexibility for meaningful interaction with patients.

The emergence of large language models (LLMs) offers new possibilities for rapidly building follow-up dialogue systems. LLMs are self-supervised models trained on massive text (and multimodal) data, capable of understanding, generating, reasoning, and interacting through language [12]. In medical follow-up scenarios, LLM-based systems require neither complex rule sets nor large annotated datasets, reducing development and maintenance costs. Their strong language capabilities enable more natural and humanized conversations, improving patient empathy and engagement[15]. Therefore, LLM-based follow-up chatbots show both technical feasibility and potential in enhancing patient adherence and care. Chen et al.[3] developed a follow-up chatbot for postoperative management with report generation and flow adaptation but lacked support for complex forms. Yang et al. [20] proposed Talk2Care for elderly users, supporting natural conversation and summarization to improve home follow-up. Tu et al. [16] introduced TRUST for PTSD assessment via LLM-driven clinical interview simulation, achieving expert-level diagnosis. Wei et al. [18] showed that prompt design significantly affects the performance of self-report data collection.

Directly inputting medical follow-up forms into LLMs to automatically generate questions, extract answers, and store data is a straightforward approach to building follow-up chatbots. However, our investigation into follow-up forms for the ten most common cancers revealed significant diversity and structural complexity—varying by disease type, clinical stage, and patient profile, and often involving conditional jumps, nested logic, and mixed question types. In real-world long-context, multi-turn interactions, this leads to information loss or hallucinations [9, 10], making accurate follow-up challenging. Moreover, end-to-end generation struggles to generalize across varied forms and fails to capture hierarchical and logical dependencies, reducing controllability and reliability [8].

Ensuring both flexibility and structured control thus remains a key challenge in applying LLMs to medical follow-up.

To investigate the limitations of applying large language models in an end-to-end manner for medical follow-up chatbots, we conducted experiments based on prior research on follow-up and patient data collection systems, as well as forms obtained from clinical experts. We selected three types of follow-up forms for comparison: (1) simple forms with a small number of questions (10, single choice), (2) medium-complexity forms with more questions (45, single choice) but no logic jumps or complex types, and (3) high-complexity forms (53 questions) that include multiple question types (single choice, multiple choice, and fill-in-the-blank) along with nested and conditional logic.

By comparing an end-to-end LLM-based approach with our proposed modular follow-up strategy, we analyze performance across different levels of form complexity and systematically identify key limitations of LLMs in real-world follow-up scenarios.

The main contributions of this work are as follows:
- We conduct comparative experiments across follow-up forms with varying complexity (in terms of question volume, types, and branching logic) to analyze the performance of large language models in question generation, context retention, option recognition, and logic handling. Our results systematically reveal key limitations of end-to-end approaches in complex follow-up scenarios, including repeated questioning, premature termination, and failure to produce structured outputs.
- To address these issues, we propose a form-aware modular control framework that orchestrates question generation, answer extraction, and logic transitions. This structured mechanism ensures the integrity of the dialogue flow and the accuracy of structured outputs, thereby enhancing both the flexibility and the practical deployability of LLM-based follow-up systems..

Our implementation, prompts, and demo examples are available at https://github.com/LiuJnYn/LLM_follow-up

## 2 METHODS

Our main research methodology is illustrated in Figure 1. First, we analyzed all the follow-up forms currently obtained from discussions with clinical informatics experts to understand real-world follow-up needs. Second, based on Qwen-plus, we utilized prompt engineering to set up three distinct virtual patients to simulate real-world conversations between patients and the follow-up dialogue chatbots. Next, we established an end-to-end follow-up dialogue robot as the control group and a modularly controlled follow-up dialogue robot as the experimental group.

### 2.1 Follow-up Form Analysis

Based on discussions with clinical informatics experts and the analysis of follow-up forms collected from real-world medical projects, we obtained a total of 32 follow-up forms. We examined key parameters such as the number of questions, question types, and the presence of logical branching. Our analysis revealed that the number of questions per form ranges from 10 to 146. Over half of the follow-up forms fall within the 10–50 question range, reflecting a concentrated distribution around this interval. Question types include single-choice, multiple-choice, free-text, mixed choice-text input, checklist, and body map formats. Additionally, over 50% of the forms involve logical branching between questions, indicating that handling branching logic is an essential and unavoidable aspect of medical follow-up forms.

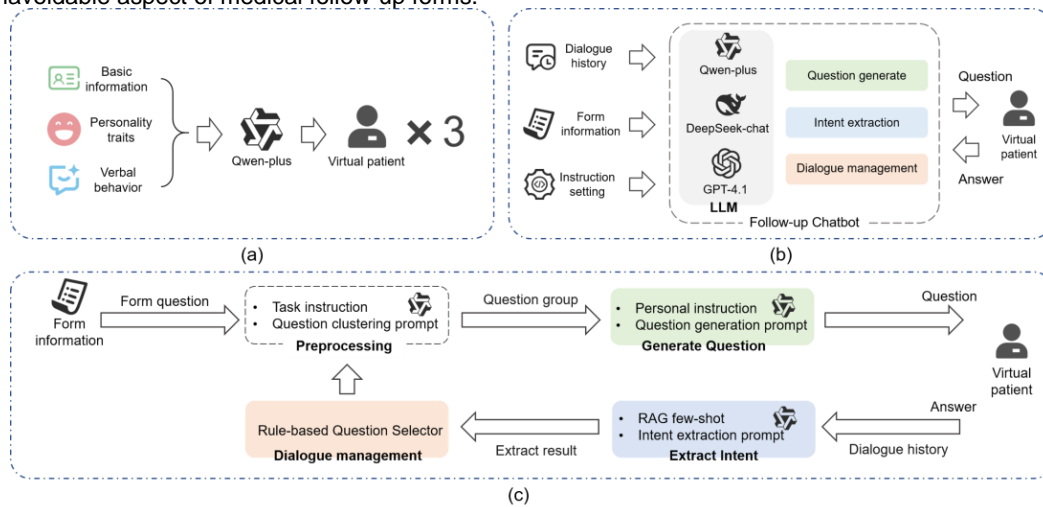

Figure 1.Overall Architecture of the Follow-up Dialogue System and Experimental Framework. (a) Three virtual patients were constructed by inputting their basic information, personality traits, and few-shot dialogue examples into the language model to simulate distinct verbal behavior. (b) In the control group, an end-to-end follow-up chatbot was configured by prompting the language model with dialogue history, form information, and instructions, allowing the model to autonomously handle question generation, intent extraction, and dialogue flow management. We test three LLM. (c) In the experimental group, a programmatically controlled follow-up chatbot was implemented by decomposing key tasks and independently designing modules for question generation, intent extraction, and dialogue management.

## 2.2 Follow-up Chatbot Designs

### 2.2.1 Virtual Patient Design

To simulate real-world interactions between patients and follow-up dialogue agents, we designed structured but content-varied virtual patient prompts to guide the Qwen-plus[19] language model in role-playing virtual patients. These virtual patients engaged separately with both the end-to-end and the flow-controlled follow-up chatbots. As illustrated in Figure 1(a), each prompt consisted of three components. First, we defined personal background information—such as name, age, occupation, and residence—to establish the patient's identity. Second, we assigned distinct profiles for each of the three virtual patients, in order to introduce diversity in language style and behavioral responses. Finally, we extracted and revised several turns of real follow-up conversations from previously collected data and used them as few-shot examples aligned with each virtual persona. These examples helped guide the model to generate contextually coherent and realistic responses, enabling a comprehensive evaluation of the chatbot's stability and adaptability across diverse patient types.

### 2.2.2 End-to-End LLM-based Chatbot

As shown in Figure 1(b), we designed an end-to-end follow-up dialogue system based on a large language model (LLM) as the control group. The system is driven by a natural language prompt, which consists of three components: 1) Form information: the three follow-up forms introduced in Section 3.1 are transformed into a unified textual structure that includes the full questionnaire content, each question's options, and their skip and nested logic, to guide question generation and dialogue progression; 2) Dialogue history: the completed dialogue rounds are provided to help the model track the dialogue state within context; 3) Instruction setting: a system prompt and procedural task chain instruct the model to read and understand the form and dialogue history, perform question generation, answer extraction, and logical reasoning, thus constraining the model's behavior to ensure the dialogue progresses appropriately.

Based on this prompt, the model must independently perform question generation, intent extraction, and dialogue flow control without any external programmatic guidance, and output a structured JSON result. We selected three mainstream general-purpose LLMs—Qwen-plus, DeepSeek-chat[5], and GPT-4.1[13]—as base models. Under identical input conditions, each virtual patient interacted with each model on each follow-up form through 10 rounds of simulated Q&A to evaluate performance in handling long forms, multi-turn understanding, and dialogue management.

*2.2.3 Modular LLM-based Chatbot*

To enhance the reliability and controllability of the follow-up dialogue process, we designed a modular LLM-based follow-up chatbot using Qwen-plus, which serves as the experimental group. This system comprises five main modules: form structure construction, question preprocessing, question generation, dialogue information extraction, and dialogue flow control.

In the form structure construction phase, we converted the original follow-up forms into structured JSON format. Each entry includes the question text, question type, options for single/multiple-choice questions, suffix text for fill-in-the-blank items, and conditional logic for branching.

Next, during the preprocessing phase, the form questions were categorized by type (single-choice, multiple-choice, fill-in-the-blank), and questions of the same type were input into the LLM for semantic clustering. This design mirrors real-world follow-up practices observed in actual recordings, where interviewers often combine related questions into a single inquiry to improve communication efficiency and naturalness. Specifically, we first constructed an abstraction prompt to instruct the LLM to summarize and analyze all questions, producing a content-level description. Then, this summary and the original questions were jointly input into a clustering prompt, guiding the model to group semantically or structurally similar questions into clusters. Each form underwent multiple clustering trials, and the most frequently occurring result was selected as the final grouping.

For each question group, we designed distinct question generation prompts according to the question type, enabling the model to generate human-like, conversational queries. This approach addresses patients' potential difficulty in understanding highly clinical or technical phrasing in the original forms [14].

After receiving a response from the virtual patient, we combined the form entry and current dialogue context and input them into the LLM along with an intent extraction prompt, tailored to each question type. We designed a RAG-based method for intent extraction to enhance the large language model's ability to extract information based on the current dialogue context. The construction process for the RAG knowledge base is as follows: After the LLM generates natural language questions from the form content, we randomly select a preset user intent corresponding to the dialogue and instruct the three virtual patients to provide a natural language response based on that intent. This process automatically generates a "question - patient response - intent extraction result" dataset. During the intent extraction process, we retrieve the dialogues from the knowledge base that are most similar to the current conversation. These are used as few-shot examples to prompt the large language model with the extraction results from these similar dialogues. LLM was instructed to extract the correct structured information corresponding to the form item from the user's response.

The question selection module then determines the next group of questions based on three rules:

(1) If any items in the current question group fail to yield extracted intent, the unanswered items are regrouped and asked again.

(2) If all items have valid intent extracted, the system checks whether any responses trigger follow-up questions (e.g., if the patient answers "quit smoking" to "Do you smoke?", additional questions about quitting are required). If so, the follow-up questions are returned to the clustering module and re-enter the full processing loop.

(3) If no new questions are triggered, the system proceeds to the next predefined group of questions.

## 3 RESULT

### 3.1 Form Selection Results

Based on the statistical analysis result, we selected three representative forms to evaluate the capabilities of large language models. Form-1 is a simple form containing a small number of questions (10 questions, all single-choice), primarily used to assess how post-treatment pain affects patients' daily lives. Form-2 is a medium-complexity form with a larger number of questions (45 questions, also single-choice) focused on evaluating patients' quality of life. Form-3 is a high-complexity form consisting of 53 questions that include single-choice, multiple-choice, and fill-in-the-blank types, along with embedded skip logic. It is designed to collect comprehensive information on patients' overall health status.

### 3.2 Evaluation of Follow-up Chatbot Systems

By combining demographic information, personality traits, and real few-shot dialogue examples, we successfully constructed three types of virtual patients to reflect common response styles in real-world follow-up scenarios: clear and concise (patient-1), clear but verbose (patient-2), and vague or off-topic (patient-3). The benefit of this design is that it transcends a simple "correct answer" test, providing a more challenging and realistic benchmark for evaluating the robustness and adaptability of dialogue systems.

Through systematic analysis of the dialogue generation process in the control group, we observed that all tested LLMs were able to complete the follow-up tasks for Form-1 with high accuracy in flow control and intent extraction. However, when handling longer or more complex forms involving skip logic, various issues emerged. Some models failed to initiate the conversation from the first question, leading to deviations from the intended dialogue flow. Others prematurely terminated the interaction before completing all questions, resulting in incomplete information collection. Additionally, excessively long response times negatively impacted interaction efficiency. In content generation, certain models altered the wording or structure of the original questions—an issue most frequently observed with patient-3. Other common problems included repeated questions, errors in logic jumps, and skipped or missed items. These findings highlight the limitations of the end-to-end approach in managing structured questionnaires, handling complex skip logic, and enforcing task constraints. A detailed classification of error types and their frequency across different models and forms is provided in Table 1.

Table 1: Summary of Error Types Presented in Form-2 and Form-3

| Form Type | Error Type | Qwen-Plus | DeepSeek-Chat | GPT-4.1 |
|---|---|---|---|---|
| Form-2 | Starting from the middle | 0 | 0 | 1 |
|  | Ending dialogue prematurely | 2 | 0 | 0 |
|  | Excessive response time | 1 | 11 | 5 |
|  | Altering questions | 3 | 8 | 8 |
|  | Repetitive questioning | 19 | 16 | 29 |
|  | Logical jump errors | \ | \ | \ |
|  | Skipping/Missing questions | 6 | 4 | 1 |
| Form-3 | Starting from the middle | 3 | 2 | 0 |
|  | Ending dialogue prematurely | 20 | 3 | 5 |
|  | Excessive response time | 4 | 9 | 1 |
|  | Altering questions | 12 | 13 | 18 |
|  | Repetitive questioning | 11 | 23 | 15 |
|  | Logical jump errors | 10 | 8 | 5 |
|  | Skipping/Missing questions | 13 | 7 | 5 |

We also manually evaluated the intent extraction accuracy of different large language models across various forms and virtual patient profiles. Results show that under the experimental setting of Form-2, the models initially maintained a structured mapping of intents with relatively high accuracy. However, as the dialogue progressed, the error rate increased, and issues emerged where responses were not aligned with the predefined options. In the more complex Form-3, extraction errors were primarily due to the models failing to structure free-text responses from fill-in-the-blank questions—instead copying the virtual patient's full sentence without parsing it. A comparison of intent extraction accuracy across models is illustrated in Figure 2, where we observe a consistent decline in performance as the form length and complexity increase. To enhance the visibility of performance difference, the y-axis is intentionally truncated to start at 0.5.

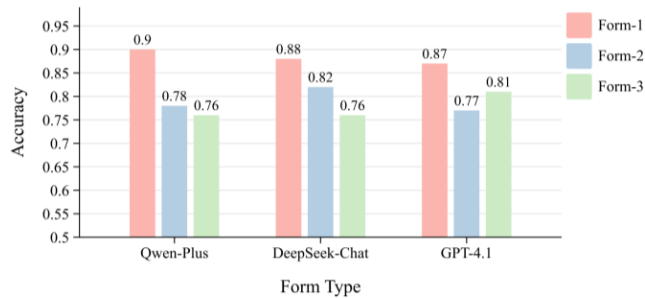

Figure 2. Comparison of Intent Extraction Accuracy in End-to-End Follow-up Chatbot. Note that the Y-axis (Accuracy) starts from 0.5.

In the experimental setup, the semantic clustering module powered by a large language model automatically grouped semantically similar items within the form and merged them into composite questions, enabling more efficient and natural follow-up interactions. Compared to the end-to-end chatbot, the average number of dialogue turns was reduced by 64% for Form-1, 48.1% for Form-2, and 28.1% for Form-3. It is worth noting that for Form-2, the average number of turns only reflects results from Qwen-Plus and DeepSeek-Chat, as GPT-4.1 failed to complete form completion within the maximum allowed 80 turns in all test cases. Overall, the average reduction in dialogue turns across all forms was 46.73%, significantly improving information collection efficiency.

Figure 3 compares the average response time, token consumption, and intent extraction accuracy between the experimental and control methods using the Qwen-Plus model. The slightly longer average response time in the control group (+0.47 seconds) is attributed to longer context lengths that the model needs to process. While the difference is not large, the control group occasionally experienced timeout failures due to excessively long responses (see Table 1), which did not occur in the experimental group.

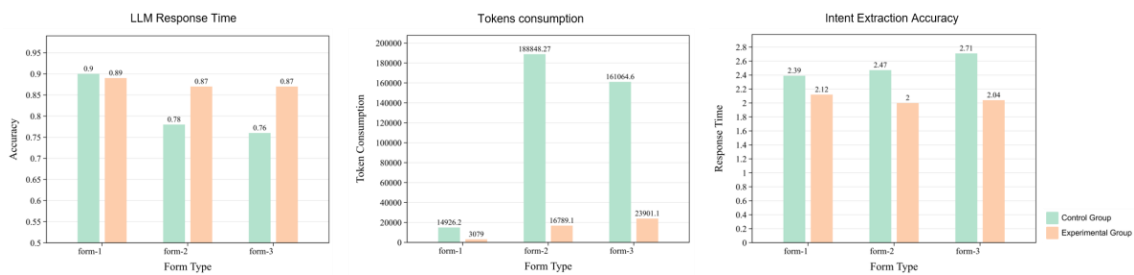

Figure 3. Performance Comparison Between Experimental and Control Chatbots Across Key Metrics. Note that the LLM Response Time Y-axis (Accuracy) starts from 0.5.

In terms of system efficiency, the token consumption per completed turn in the control group was 5–8 times higher than that of the experimental group, and unexpected behavior was also observed. When dealing with shorter forms, both groups achieved intent extraction accuracy around 0.9, demonstrating the language model's capability in understanding and extracting intent. However, as the form length and complexity increased (in Form-2 and Form-3), the control group's accuracy dropped significantly—mainly due to poor handling of frequency-related items, unstructured extraction from fill-in-the-blank answers, and failure to adhere to predefined option sets. In contrast, the experimental group consistently maintained high and stable accuracy across conditions.

## 4 DISCUSSION

This study systematically compares the performance differences between end-to-end and modular LLM-driven follow-up dialogue systems under varying form complexities, revealing the core challenges and potential optimization directions for current large models in medical follow-up applications.

First, the experimental results indicate that while the end-to-end method possesses basic usability for short forms, it is prone to issues such as dialogue flow interruption, repetitive questioning, and skipping or missing questions when faced with a large volume of items and high logical complexity. This primarily stems from the LLM's deficiencies in long-context retention, logical reasoning, and task-boundary awareness. Lacking external structural constraints, the model easily falls into infinite generation loops or transition failures. This is especially true in intent extraction tasks, where accuracy significantly declines with increasing context length, severely impacting system stability and reliability. Furthermore, the token consumption of the control group models on complex tasks was 5–8 times that of the experimental group, with extreme resource waste observed, such as one session reaching 3,133 turns and consuming nearly 200 million tokens.

In contrast, our proposed modular approach—by leveraging form structure parsing, semantic clustering, question generation, intent extraction, and a rule-driven flow scheduling mechanism—can effectively guarantee the completeness of task progression and the structured nature of the output. It also retains the LLM's natural language understanding and generation advantages in terms of interaction flexibility and patient adaptability. This method not only reduces the number of interaction turns by 46.73% but also demonstrates higher accuracy and robustness in handling complex logic and information extraction, all while preserving the LLM's natural conversational ability, thus possessing stronger deployment feasibility.

Despite this, our method still relies on rule and prompt design and lacks high-level generalizability. Although the virtual patient setup enhanced the controllability of the experiment, there is still a gap compared to interactions with real patients. Future research can further explore adaptive flow control mechanisms, patient-feature-driven personalization strategies, and multimodal follow-up systems to promote the safe, controllable, and efficient application of LLMs in medical follow-up.

## 5 CONCLUSION

This study focuses on exploring the limitations of large language models in real-world medical settings and compares the performance of end-to-end and modular LLM-based follow-up dialogue systems in interactions with virtual patients. Experimental results show that as form complexity increases, end-to-end methods face challenges such as uncontrollable dialogue flow, inaccurate intent extraction, and excessive resource consumption. To address these issues, we propose a modular approach that decomposes the follow-up task and incorporates semantic clustering and flow control mechanisms. This method not only effectively completes the follow-up process but also significantly reduces the number of dialogue turns and token consumption. Our findings highlight that LLMs alone are insufficient for handling complex tasks in highly structured and accuracy-critical medical scenarios. External control mechanisms and carefully designed workflows are essential. Future work may explore integrating personalized patient information to enhance user experience, as well as developing more automated dialogue management strategies to advance intelligent follow-up systems toward greater safety, personalization, controllability, and efficiency. Furthermore, the current method remains based on paper forms and is limited to text-based interaction. Future research should explore multimodal capabilities to accommodate the demands of more complex medical follow-up scenarios.

## ACKNOWLEDGMENTS

This work was supported by the Research and Development Project of "Jianbing" "Lingyan" of Zhejiang Province [grant numbers 2025C01128]